\documentclass[a4paper,UKenglish,cleveref, autoref]{lipics-v2019}

\usepackage{caption}
\usepackage{subcaption}
\usepackage{adjustbox}
\usepackage{amsmath}
\usepackage[ruled,vlined]{algorithm2e}

\title{Traffic Prediction Framework for OpenStreetMap using Deep Learning based Complex Event Processing and Open Traffic Cameras} 

\titlerunning{Traffic Prediction Framework for OSM using Open Traffic Cameras}

\author{Piyush Yadav\footnote{corresponding author}}{Lero-SFI Irish Software Research Centre, Data Science Institute \and National University of Ireland Galway (NUI Galway), Ireland }{piyush.yadav@lero.ie}{https://orcid.org/0000-0002-4872-0205}{}

\author{Dipto Sarkar}{Department of Geography, University College Cork, Ireland}{dipto.sarkar@ucc.ie}{https://orcid.org/0000-0003-2254-049X}{}

\author{Dhaval Salwala}{Insight Centre for Data Analytics, NUI Galway, Ireland}{}{}{}

\author{Edward Curry}{Insight Centre for Data Analytics, NUI Galway,  Ireland}{edward.curry@insight-centre.org}{https://orcid.org/0000-0001-8236-6433}{}

\authorrunning{P.Yadav, D. Sarkar, D.Salwala and E.Curry}

\Copyright{Piyush Yadav, Dipto Sarkar, Dhaval Salwala and Edward Curry}



\relatedversion{}

\supplement{}

\funding{This work was supported by the Science Foundation Ireland grants SFI/13/RC/2094 and SFI/12/RC/2289\_P2.}

\nolinenumbers 

\EventEditors{}
\EventNoEds{2}
\EventLongTitle{11th International Conference on Geographic Information Science (GIScience 2021)}
\EventShortTitle{GIScience 2021}
\EventAcronym{GIScience}
\EventYear{2021}
\EventDate{September 27--30, 2021}
\EventLogo{}
\SeriesVolume{177}
\ArticleNo{17}

\pdfoutput=1
\begin{document}
\maketitle
\begin{abstract}
Displaying near-real-time traffic information is a useful feature of digital navigation maps. However, most commercial providers rely on privacy-compromising measures such as deriving location information from cellphones to estimate traffic. The lack of an open-source traffic estimation method using open data platforms is a bottleneck for building sophisticated navigation services on top of OpenStreetMap (OSM). We propose a deep learning-based Complex Event Processing (CEP) method that relies on publicly available video camera streams for traffic estimation. The proposed framework performs near-real-time object detection and objects property extraction across camera clusters in parallel to derive multiple measures related to traffic with the results visualized on OpenStreetMap. The estimation of object properties (e.g. vehicle speed, count, direction) provides multidimensional data that can be leveraged to create metrics and visualization for congestion beyond commonly used density-based measures. Our approach couples both flow and count measures during interpolation by considering each vehicle as a sample point and their speed as weight. We demonstrate multidimensional traffic metrics (e.g. flow rate, congestion estimation) over OSM by processing 22 traffic cameras from London streets. The system achieves a near-real-time performance of 1.42 seconds median latency and an average F-score of 0.80.

\end{abstract}

\section{Introduction}
\label{sec:Introduction}
OpenStreetMap (OSM) is arguably the largest crowdsourced geographic database. Currently, there are more than 5 million registered users,  over 1 million of whom have contributed data by editing the map. Different aspects of OSM data quality have been scrutinized, and OSM data performed well on tests of volume, completeness, and accuracy across several classes of spatial data, such as roads and buildings ~\cite{neis2012street,haklay2010good}. Despite standing up to data quality tests, enforcing data integrity rules required for a navigable road map is challenging. Lack of topological integrity and semantic rules such as turn restrictions which enable navigational capabilities is a stumbling block for OSM to be a viable open-source alternative to commercial products like Google Maps and Apple Maps.

Over the last five years, several large corporations have realized the value of OSM and have assembled teams to contribute and improve data on the OSM platform ~\cite{anderson2019corporate}. Backed by large companies, the editing teams are capable of editing millions of kilometers (km) of road data each year. As a result of these efforts, the road network data on OSM is improving rapidly, and the gap in the quality of the data in the developed and developing countries is narrowing. In the foreseeable future, it is expected that widely available open-source navigation services may also be built on top of OSM data. The next frontier to making OSM more usable for navigational purposes is to have real-time traffic information to estimate trip times. This feature is already available in commercial digital navigation maps. However, the estimation and availability of traffic state data relies on platforms (i.e. iPhone and Android) built by the respective companies to feed data to the service.

In recent years, internet-connected devices collectively referred to as the Internet of Things (IoT) have become ubiquitous. With the proliferation of visual sensors, there is now a significant shift in the data landscape. We are currently transitioning to an era of the Internet of Multimedia Things (IoMT) ~\cite{alvi2015internet,aslam2018towards} where media capturing sensors produce streaming data from different sources like CCTV cameras, smartphones and social media platforms. For example, cities like London, Beijing and New York have deployed thousands of CCTV cameras streaming hours of videos daily ~\cite{somebody}. Complex Event Processing (CEP) is an event-driven paradigm which utilizes low-level data from sensor streams to gain high-level insights and event patterns. CEP applications can be found in areas as varied as environmental monitoring to stock market analysis ~\cite{demers2006towards}.

In this research, we propose a Complex Event Processing (CEP) framework to estimate traffic using deep learning techniques on publicly available street camera video streams. We process the data on GPU computing infrastructure and expose the processed output to OSM. The proposed method utilizes publicly available data and does not piggy-back on using cell phones. The solution is better for privacy and is also immune to subversion techniques such as the recent case where an artist used 99 Android phones to simulate traffic on Google Maps ~\cite{googlehack}. Further, the rich data stream from the video can provide multidimensional measurements of traffic state going beyond simple vehicle count-based metrics.

\section{Background and Related Work}

This section provides the initial background and techniques required for the development of a traffic estimation service for OSM.

\subsection{Complex Event Processing}

Complex Event Processing (CEP) is a specialized information flow processing (IFP) technique which detects patterns over incoming data streams ~\cite{cugola2012processing}. CEP systems receive data from different sensor streams and then mine high-level patterns in real-time to notify users. CEP systems perform online and offline computations and can handle high volume and variety of data. In CEP, event patterns of interest are expressed using SQL like query language, and once a query is registered, it continuously monitors the data stream. Pattern matching for queried patterns occurs in (near) real-time as the data flows from the Data Producers (sensors) to the Consumers (applications).

\begin{figure}
\centering
\begin{minipage}[t]{0.5\textwidth}
  \centering
  \includegraphics[height=3.1cm, width=6.8cm]{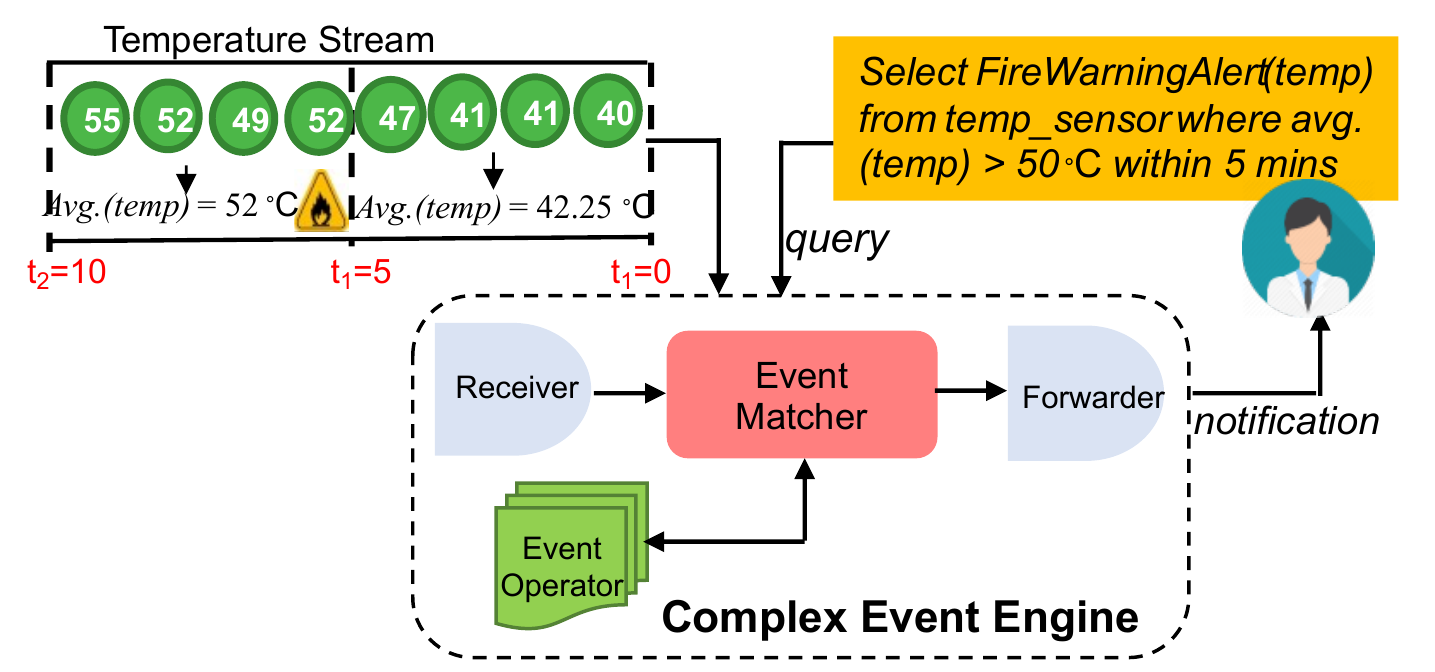}
  \caption{Complex Event Processing Paradigm. Temperature is being monitored from a sensor to issue fire warnings.}
  \label{fig:CEP}
\end{minipage}%
\hfill
\begin{minipage}[t]{0.48\textwidth}
  \centering
  \includegraphics[height=2.5cm, width=6.5cm]{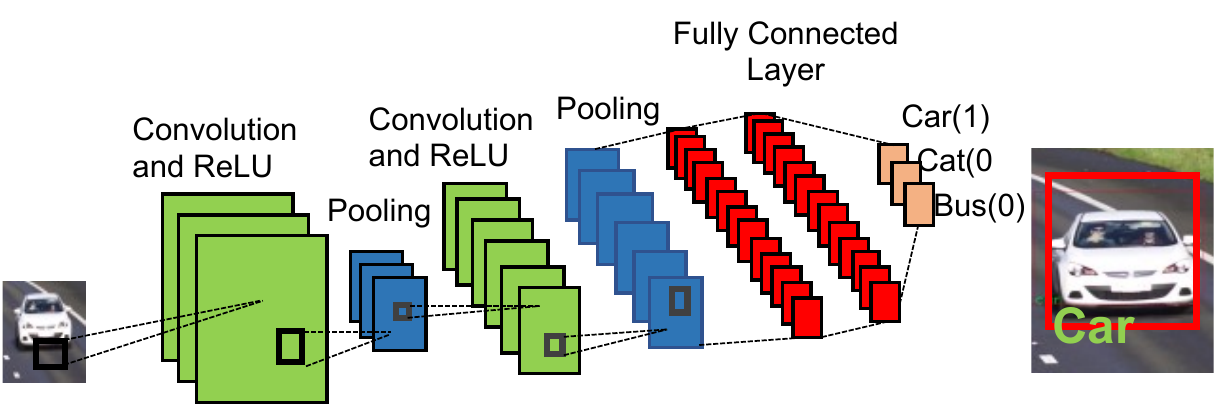}
         \caption{CNN Architecture. A CNN consisting of several connected layers of neuron for object identification from an image.}
         \label{fig:CNNA}
\end{minipage}
\end{figure}

Fig. \ref{fig:CEP} shows a simple CEP system, where a user queries from a temperature sensor to send notification of a \textit{fire warning} alert if the average temperature is greater than 50°C in the last five minutes ~\cite{yadav2019vekg}. The fire warning alert query is registered in the CEP engine which continuously monitors the data from the temperature sensor. The CEP engine will raise a fire warning alert at time $t_1-t_2$ as the average temperature of incoming streams is higher than 50°C in the last five minutes. Thus, a complex fire warning alert is generated by averaging simple ‘temperature event’ from the sensor.

\subsection{Deep Learning-based Image Understanding}

The computer vision domain focuses on reasoning and identifying image content in terms of high-level semantic concepts. These high-level concepts are termed as \textit{objects} (e.g. car, person) which act as building blocks in understanding and querying an image. There are different object detection algorithms like SIFT ~\cite{lowe1999object}, which classify and localize the objects in the video frames. Convolutional Neural Networks (CNN) based deep learning methods ~\cite{lecun2015deep} are proficient at identifying objects with high accuracy. It is a supervised learning technique where layers are trained using labelled datasets to classify images.  Fig. \ref{fig:CNNA} shows the underlying architecture of a CNN model having different layers to classify and detect an object from the image. CNN based object detection methods like YOLO ~\cite{redmon2016you} detect and classify objects by drawing bounding boxes around them.

\subsection{Related Work}

\textbf{Traffic Estimation Services}: Most of the traffic monitoring related data is collected from sensor devices like GPS embedded in mobile phones and speed detection cameras (loop detectors, camera, infrared detectors, ultrasonic, radar detectors) ~\cite{andreas2009traffic}. The traffic data is displayed on proprietary maps such as Google, Apple and Here Maps. These companies also expose the data through APIs. However, the methodology for traffic estimation is opaque, and depending on the service provider, there are costs and restrictions on how a third-party developer can utilize the data. Initiatives such as OpenTraffic\footnote{\url{opentraffic.io}} are building an open-source data platform and services which collect anonymized telemetry data from vehicles and smartphones and exposes them through OSM API. OpenTraffic is a relatively new platform and is still building up a list of partners (currently three) to gather traffic information. Other attempts to create open traffic data framework for consumption as a map service include the OpenTransportMap\footnote{\url{opentransportmap.info}} project. Their framework is based on pre-calculated estimates of traffic volumes based on demographic data. There are some traffic state prediction works which use interpolation techniques using sensor-based reading like GPS and LIDAR to identify traffic at unknown locations ~\cite{lowry2014spatial,zou2012improved,song2018traffic}. In this work, instead of telemetry data like GPS, we focused on clusters of openly available video feeds which provide live-streaming updates from multiple locations and update traffic in real-time on OSM.

\noindent \textbf{Video-Based Traffic Estimation}: Video streams are an ideal example of BigData as it represents a high volume, high velocity, unstructured source of data. Fatih et al. ~\cite{porikli2004traffic} proposed a Gaussian based Hidden Markov Model (GM-HMM) to estimate traffic over MPEG videos. But their work was limited only to the camera Field of View (FoV) to determine the traffic over a highway segment. In this work, we focused on estimating traffic beyond camera FoV across the whole queried street network even where camera feeds are not available. Kopsiaftis et al. ~\cite{kopsiaftis2015vehicle} used background estimation over high-resolution satellite video images and performed traffic density estimation by counting vehicles in the given region. Again, their work is limited to only pre-recorded historical video data. On the other hand, our proposed framework estimates traffic over streaming video in near-real-time. Connected vehicles are another data source for traffic estimation. Kar et al. ~\cite{kar2017real} proposed real-time traffic estimation considering vehicles as edge nodes. They performed object and lane detection using dash cameras installed in the vehicles to estimate their speed. The author's future work was focused on sharing such data across vehicles and on a cloud-based map service. Our work builds upon this line of argument by considering traffic camera network as a cluster of edge nodes and then applying data-driven techniques to identify traffic across the road network which is later updated to OSM.

\begin{figure}
\centering
\begin{minipage}[t]{.5\textwidth}
  \centering
  \includegraphics[height=3.8cm, width=7cm]{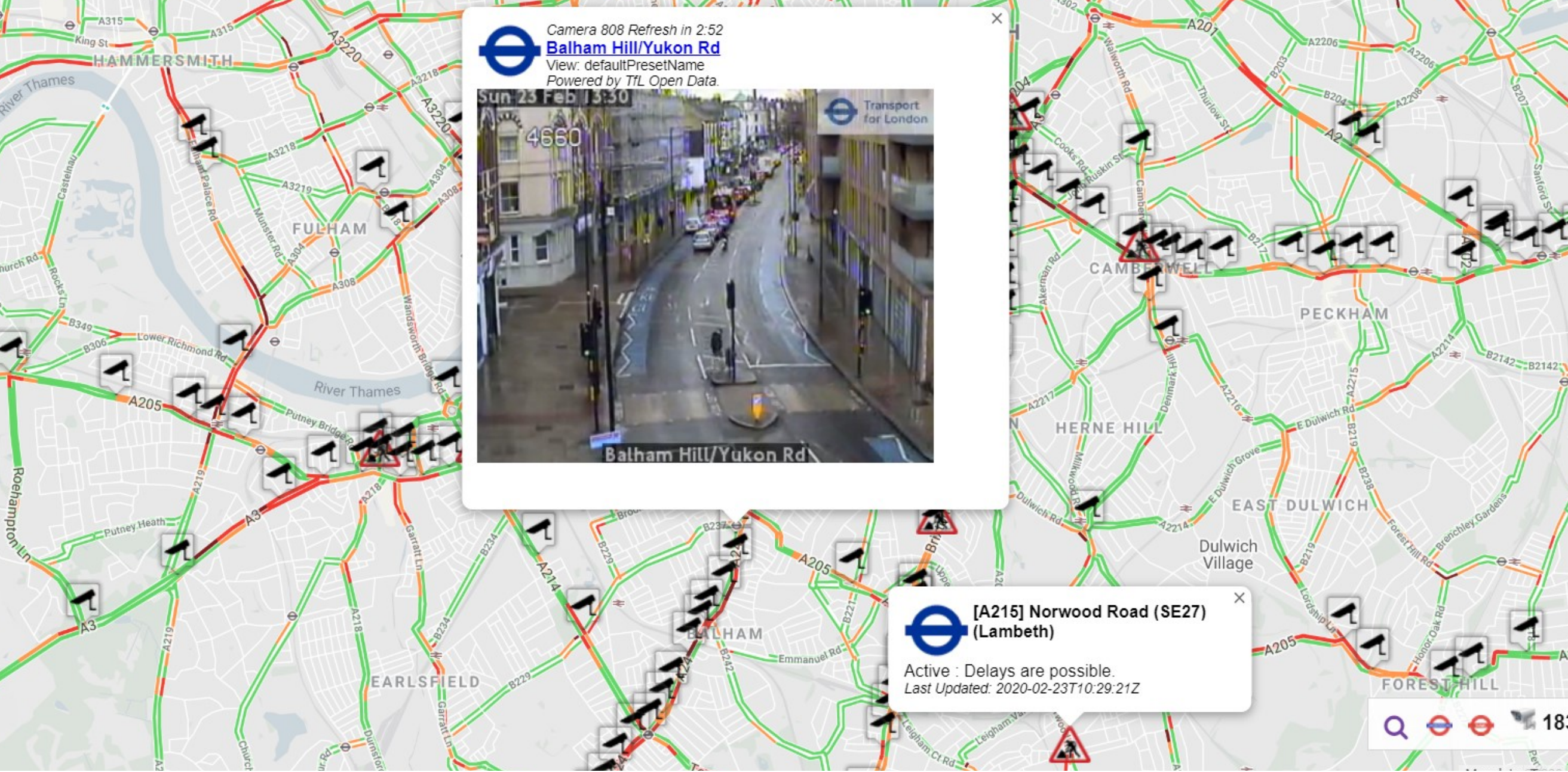}
  \captionof{figure}{TfL Traffic Camera Dashboard}
  \label{fig:TfL}
\end{minipage}%
\begin{minipage}[t]{.5\textwidth}
  \centering
  \includegraphics[height=3.8cm, width=7cm]{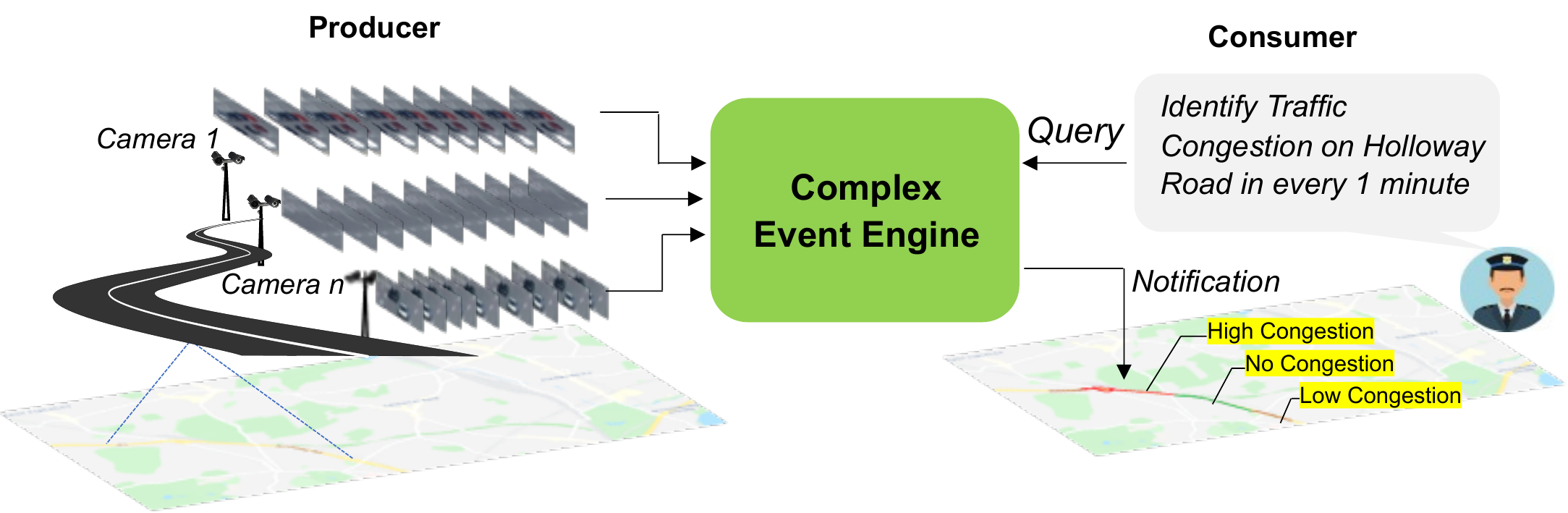}
  \captionof{figure}{Traffic CEP Overview}
  \label{fig:CEPTraffic}
\end{minipage}
\end{figure}

\section{Motivation and Problem Scope}

\subsection{Case Study: London- A City of Open Traffic Cameras}

London possesses an extensive camera network. The city is dotted with nearly 500K cameras ~\cite{londoncam}. With a density of 68.4 cameras per 1000 people, it is estimated that an average Londoner is caught in a camera approx 300 times a day ~\cite{londoncam}. Live camera feeds stream real-time information of things happening around London's streets, enabling applications like license plate reading. The camera streams API is provided from Transport for London (TfL) and can be accessed by registering with their system. Each video camera comes with metadata including date, timestamp, street name and its geolocation. Availability of open cameras, accessibility of real time\footnote{\url{https://www.tfljamcams.net/}} and archived data\footnote{\url{http://archive.tfljamcams.net/}}, and friendly streaming API makes London ideal for our case study. Fig. \ref{fig:TfL} shows the screenshot of camera networks across a part of the city with a video clip instance from a particular camera.

\subsection{User Scenario}

Traditional traffic monitoring using CCTV is mostly a manual effort where traffic personnel monitor the situation by looking at the feeds from cameras installed across the city. This manual approach is time-consuming, tedious, and error-prone as it is difficult for humans to synthesize a large number of events occurring across space and time. Thus, the development of automated techniques is crucial to supplement manual qualitative efforts. Suppose the traffic authority wants to map the busiest routes of the city over the day in real-time and wants to provide the traffic state as a service to the citizens. Fig. \ref{fig:CEPTraffic} shows the authority using a CEP engine to obtain the traffic congestion status over a road segment from a cluster of  CCTV cameras installed along the road. Processing this query over multiple video streams requires identifying objects (e.g. cars), determining their properties (e.g. speed), calculating traffic states, interpolation across the road network for continuous estimation, and visualizing the results in real-time with summarized metrics and visualizations overlaid on OSM.

\begin{table}
\tiny
\begin{minipage}[b]{0.45\linewidth}
\begin{adjustbox}{width=8.5cm, height=2.4cm,center}
\begin{tabular}{|p{2.5cm}|p{8cm}|}
    \hline
    \textbf{ Traffic Metrics} & \textbf{ Definition} \\
    \hline
    \hline
    \textbf{Traffic Flow (TF)} & TF is the total count of vehicles that have passed a certain point in both directions for a given time. TF is calculated over short duration's (e.g. two days) and then approximated over days adjusting for seasonal and weekly variations. The TF can be calculated at different timescales ranging from minutes, hours, daily, weekly and yearly level (Annual Average Daily Traffic). \\
    \hline
        \textbf{Average Traffic Speed (ATS)} & It is the average speed of vehicles on both directions at a point for a given time scale.  ATS is measured in m/s, km/hr or mph.\\
    \hline
        \textbf{Traffic Density (TD) and Capacity(c)} & TD or \textit{volume (V)} is the number of vehicles per unit distance over a given road. It is measure in term of the number of vehicles per unit road (Km or mile). The maximum number of vehicles in a mile per lane which a road can accommodate is termed as \textit{capacity}. \\
    \hline
  \end{tabular}
  \end{adjustbox}
    \caption{Traffic Metrics}
    \label{table:1}
\end{minipage}\hfill
\begin{minipage}[b]{0.46\linewidth}
\begin{adjustbox}{width=6.3cm, height=2.0cm,center}
\centering
\begin{tabular}{|p{1.3cm}|p{5cm}|p{2cm}|}
    \hline
    \textbf{Level of Service} & \textbf{Description}  & \textbf{Volume-Capacity Ratio (V/C)} \\
    \hline
    \hline
    \textbf{A}& Free flowing and highest driving comfort. & <0.60 \\
    \hline
        \textbf{B} & Little delay and high driving comfort. & 0.60-0.70 \\
    \hline
        \textbf{C} & Some delay and acceptable level of comfort. & 0.70-0.80 \\
    \hline
        \textbf{D} & Moderate delay and some driving frustration. & 0.80-0.90 \\
    \hline
        \textbf{E} & High degree of delay and driving frustration. & 0.90-1.0 \\
    \hline
            \textbf{F} & Excessive delay and highest level of frustration. & >1.0 \\
    \hline
  \end{tabular}
  \end{adjustbox}
    \caption{Level of Service}
    \label{table:2}
\end{minipage}
\end{table}

\subsection{Multidimensional Traffic Analysis}
There are multiple factors related to traffic and untangling the impact of individual factors responsible for congestion is challenging. Congestion is a function of both: the physical way vehicles (and other road users) interact with each other, and the people’s perception of congestion (e.g. ‘the traffic is terrible today’) ~\cite{congestionmanagement}. We focus on the primary factors related to the physical dimensions of vehicular movement through the road network, namely vehicle count and average vehicle speed. Thus, at each time point, we estimate the number of vehicles at a location and the speed of each vehicle.

While these represent rudimentary aspects related to quantifying traffic state, they can be used as building blocks to more complex metrics such as traffic density, expected travel time, free flow ratio, and estimates of delay ~\cite{congestionmanagement,kopsiaftis2015vehicle}. No single indicator can be a ‘catch-all’ metric to represent the problem. Reliance on a single parameter in describing network performance paints an incomplete, or in some cases an incorrect picture of travel conditions. In this paper, we focus on building the stream data processing architecture and show its efficacy on a real-world application by focusing on calculating the basic metrics of vehicle count and vehicle speed. Table \ref{table:1} and Table \ref{table:2} lists the macroscopic traffic metrics and Level of Service(LOS) parameters ~\cite{los} which are consider for evaluation in this work.

\begin{figure}
\centering
  \includegraphics[height=4.1cm, width=7.1cm]{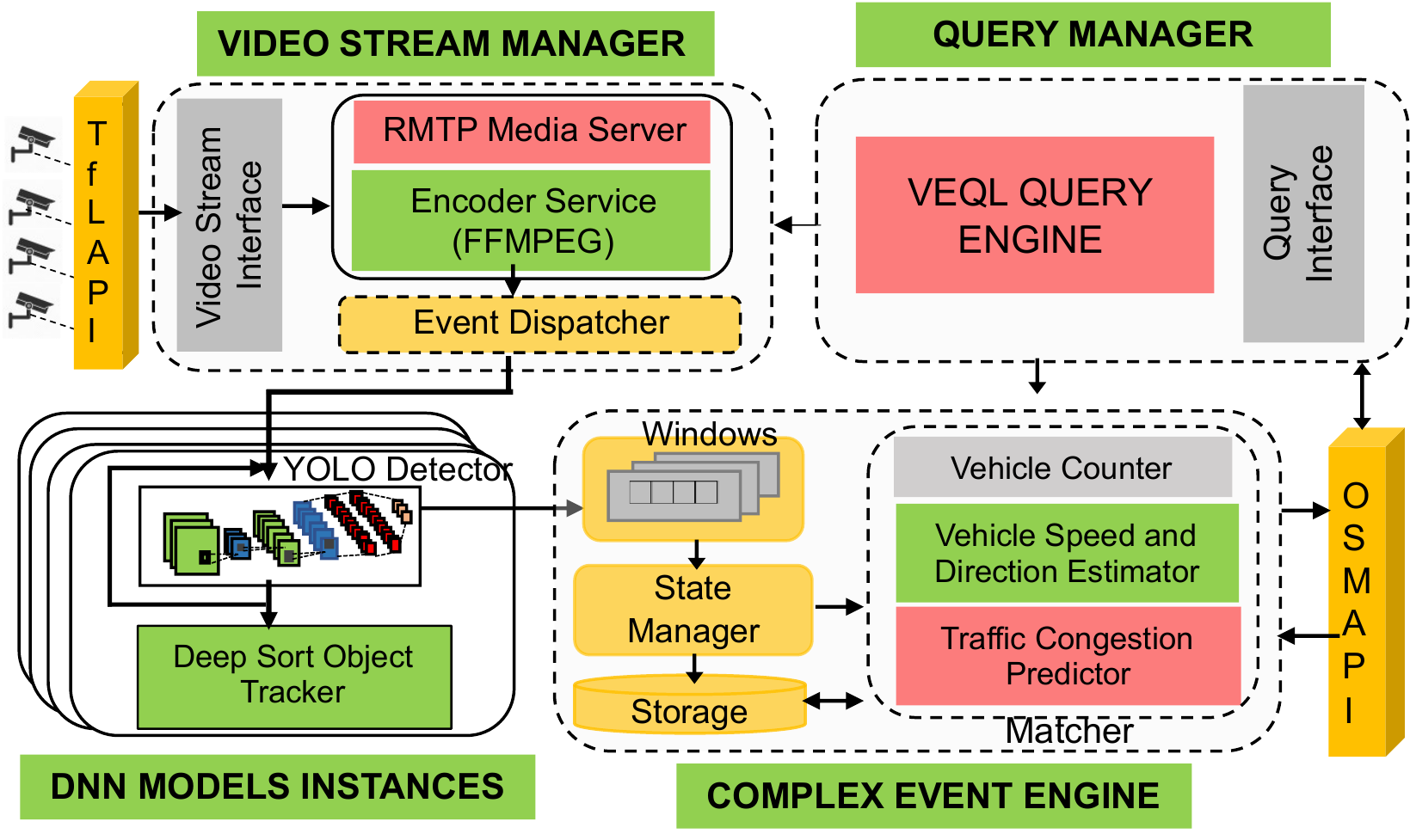}
  \caption{\centering 
  System Architecture Overview}
  \label{fig:SysArch}
 \end{figure}
 

\section{System Architecture}

To identify different traffic metrics using multiple cameras, a distributed microservice-based complex event processing system is implemented. Fig. \ref{fig:SysArch} shows the high-level architecture of the traffic estimation system which is divided into four major components. These components are independent microservices wrapped in a container and their instances can be deployed over the cloud or local computing nodes. These components are:

\begin{itemize}
    \item \textbf{Query Manager}: The user can subscribe to different queries using the query manager. The  Query manager consists of a Query Interface where users can write queries in Video Event Query Language (VEQL) to detect traffic patterns ~\cite{yadav2019vidcep}. VEQL is a SQL-like declarative language where rules and operators can be written to identify pattern over video streams. The VEQL Query Engine creates a query graph to represent patterns. Further details of VEQL can be found in ~\cite{yadav2019vidcep}. A sample VEQL query for traffic congestion is as follows:
    \begin{verbatim}
        Select Traffic_Congestion(Object) from Brixton Road
        WHERE Object = ‘Car’ OR Object = ‘Bus’
        WITHIN Time_Window = 5 sec WITH CONFIDENCE >40%
    \end{verbatim}
    In the above query, the user subscribed for Traffic Congestion for ‘Car and Bus’ over Brixton Road camera network with an update of every five-seconds. The traffic congestion operator will be discussed in detail in Section 6.
    \item \textbf{Video Stream Manager}: The video stream manager connects to multiple video feeds using a video stream interface which is a network adapter to provide the connection to different streaming API’s. The TfL (Transport for London) unified API is used to bring together video data from street cameras to our system. The Query Manager forwards the road information(e.g. Brixton Road) to the stream manager which later uses metadata information from the TfL API to fetch video streams from the corresponding road camera. For example, we pass different lat-long coordinates of Brixton Road (e.g. 51.4812,-0.11065) using OSM which is then passed to the TfL API (\url{https://api.tfl.gov.uk/Place?lat=51.4812&lon=-0.11065&radius=100&type=JamCam}) to search cameras. The search process continues and iterates until the end of the queried road segment. Similarly, each stream is sent in parallel to the media server using the \textit{GStreamer} library. Finally, the Event Dispatcher sends the received video streams from different cameras to the DNN Model pipeline.
    \item \textbf{Deep Neural Network (DNN) Models Pipelines}: This is a computer vision pipeline which consists of deep learning-based object detector and object tracker. The object detector model receives the video frames from the \textit{event dispatcher} as a feature map and extracts the vehicles in the form of bounding boxes. The object detector is integrated with a DNN based object tracker to track the identified vehicle for a given length of time. The specifics of the object detector and tracker are explained in detail in Section 5.1. Based on the number of cameras on the queried road, separate DNN model instances are created dynamically to process each camera feed in parallel. This is necessary to separate the tracking instances of each identified vehicles in each camera.
    \item \textbf{Complex Event Engine}: The Complex Event Engine is the core component which predicts traffic-related activities. Its sub-components are:
    \begin{itemize}
        \item \textbf{Window and State Manager}: Data streams (i.e. video feeds) are considered as an unbounded timestamped sequence of data [35]. CEP systems work on the concept of \textit{state}. Windows capture the stream state by taking the input stream and producing a sub-stream of finite length (equation (eq.) \ref{eq:1} and \ref{eq:2}).
        \setlength{\belowdisplayskip}{-1pt}
        \setlength{\belowdisplayshortskip}{-1pt}
        \setlength{\abovedisplayskip}{3pt}
        \setlength{\abovedisplayshortskip}{3pt}
        \begin{equation}
          S_{video} = ((f_1,t_1 ),(f_2,t_2 ),.......,(f_n,t_n ))
          \label{eq:1}
        \end{equation}
        \setlength{\belowdisplayskip}{5pt}
        \setlength{\belowdisplayshortskip}{5pt}
        \setlength{\abovedisplayskip}{0pt}
        \setlength{\abovedisplayshortskip}{0pt}
        \begin{equation}
          TIME\_WINDOW(S_{video},t_{5sec}) \colon \rightarrow S'
          \label{eq:2}
        \end{equation}

        where $f_i$ are video frames and $S^{'}$ = $((f_1,t_1 ),(f_2,t_2 ),.......,(f_n,t_{5sec} ))$. In eq. \ref{eq:2}, a $TIME\_WINDOW$ of five seconds is applied over an incoming  video stream $S_{video}$ (eq. \ref{eq:1}) and gives a  fixed sub-sequence $S^{'}$ of five seconds of video data. The \textit{State Manager} handles the created window state and sends the state information to the persistent storage and matcher services. The current work focuses on \textit{online} processing of the data in near-real-time. The \textit{Storage} component stores the event state of video feeds for historical batch-based analysis.
        \item \textbf{Matcher}: The Matcher performs traffic operations over the received state from the state manager. The matcher consists of three traffic-related operators 1) Vehicle Counter, 2) Vehicle Direction and Speed Estimator, and 3) Traffic Congestion Estimation. The functionality of these operators is explained in details in Section 5 and 6. The matcher finally sends the results of the CEP engine to the OSM layer through the OSM API.
    \end{itemize}
    
\end{itemize}

\section{Computer Vision Pipelines for Traffic Classification}
Two computer vision pipelines have been developed for the CEP Engine to estimate the traffic service. The first pipeline performs object detection (e.g. cars, buses) and tracking over incoming video streams. The second pipeline involves calculation of traffic-related properties from objects (speed, direction and count) and interpolating traffic information from point sources to create a continuous surface over OSM.

\subsection{Pipeline 1: Vehicle Detection and Tracking }
Vehicle detection is a common problem in computer vision. Different object detection techniques ranging from feature-based matching like SIFT ~\cite{lowe1999object} and complex deep learning models have been used to identify vehicles in the videos. In this work, the YOLO v3 ~\cite{redmon2016you}, a state-of-the-art object detection model is used for vehicle detection. The YOLO model considers object detection as a single regression problem and divides the image into a $S$x$S$ grid to predict the objects bounding boxes and class probabilities simultaneously.  The model gives real-time performance and processes 45 frames per second(fps) at the rate of 22 milliseconds per frame on modern GPUs and is suitable for processing streaming data like videos. Five classes of vehicles- \{\textit{bus, car, truck, bicycle and motorcycle}\} are selected as they represent the significant vehicular traffic on the road. We have used the YOLO model pre-trained on the COCO dataset which already consists of all the above five vehicular classes. The model outputs the bounding box coordinates of each detected vehicle with a probability score. The probability score is the model confidence to predict the class of an object (like vehicle), and experimentally we found that score greater than 0.4 detect most of the vehicles accurately.

Videos are timestamped continuous sequence of image frames. The vehicle object detector process frames one-by-one to detect and classify vehicles per frame. Videos are temporally correlated where the same object (vehicles) remain in multiple frames. To avoid repeatedly counting the same vehicle, the vehicle needs to be tracked across the video feed. DeepSORT \cite{wojke2017simple} is a multiobject tracking algorithm which uses Kalman filters and deep neural network to track the detected objects across the image frames. We integrated the DeepSORT tracking model with the YOLO object detection to uniquely identify each vehicle across frames.

\begin{figure}
\centering\includegraphics[width=14cm,height=3cm]{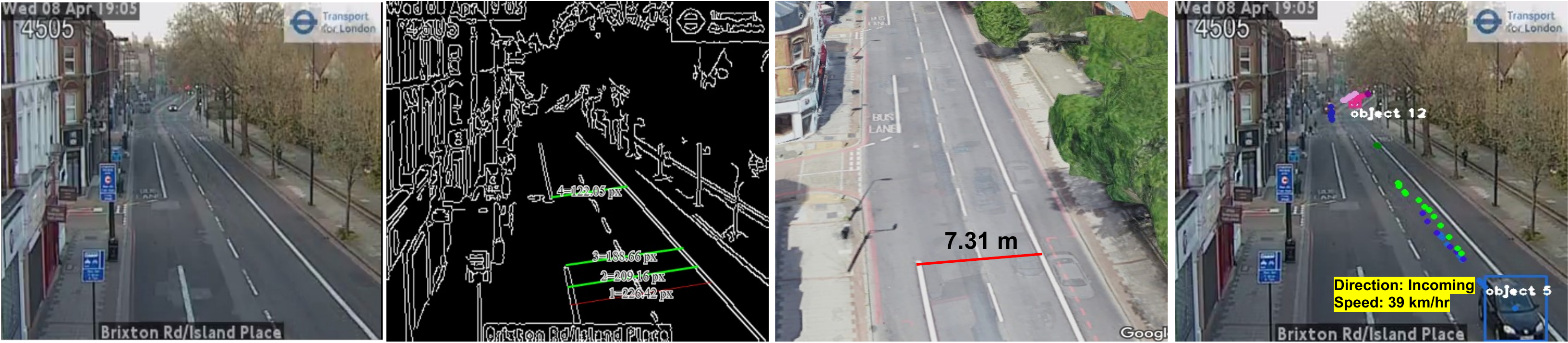}
\caption{(L-R) a)TfL Camera Image at Brixton Road/Island Place, b)Edge Detection to Identify Lanes, c) Distance Identification Using Google Earth Referencing , d) Object Identification, Tracking, Direction and Speed Estimation }
\label{fig:Googleref}
\end{figure}

\subsection{Pipeline 2: Vehicle Direction, Count and Speed Estimation}
It is essential to know the movement direction of the vehicle to segregate traffic estimation for different lanes. Direction estimation is challenging in TfL installed cameras as there is no metadata information about the direction of placement of the cameras on the road. After close inspection, we concluded that the cameras are placed over roads in such a way that their FoV covers the length of the road (top-front view). The cameras are placed in South-North or West-East direction such that outgoing traffic is in the left lane while incoming traffic is in the right lane with respect to cameras FoV. The direction of each vehicle can be calculated by measuring the displacement of the centre pixel location of its bounding box across frames for a given time window (eq. \ref{eq:2}). Considering the bottom left of the image frame as reference origin, if the displacement of the y-axis value of centre point of bounding box decreases for a given time, then it is considered an incoming traffic and vice-versa. DeepSORT provides each vehicle with a unique id so that the counting can be done for vehicles in both directions.

The speed of the vehicles can be identified using the standard equation of $Distance = Speed * Time$. Thus, the displacement of the centre pixel point of vehicle across frames where it is present for a given time (e.g. 5 sec) can be used to estimate the vehicle speed. But the speed of the vehicle will be calculated in pixels/sec which is not helpful in traffic estimation. As discussed earlier, there are no camera calibration points or pixel geotagging is available in the metadata. In computer vision, object distance from the camera is measured by taking pictures of the object from different angles and then fit into camera calibration equations. It is infeasible to go to every camera points to get object pictures from different angles. We performed a different approach to identify the relative pixels value in terms of metres. As shown in Fig. \ref{fig:Googleref}(b), a canny edge detection ~\cite{canny} algorithm is used to identify the lanes of the road using a background frame where no vehicles are present. The number of pixels between two lanes is then calculated by dividing the image into two parts and taking average pixels distance value between lanes to accommodate camera FoV at a different scale. Using the ruler tool available in Google Earth, the distance between lanes is calculated for the same location (figure \ref{fig:Googleref}(a,b,c)). For example, the number of pixels between lane is x and google earth distance is y metres, then the size of each pixel is x/y metre. We used the Design Manual for Roads and Bridges (DMRB) CD127 ~\cite{roadmanual} document which provides highway cross-sections and traffic lane width for trunk roads to validate that identified distance is within standard permissible limits. Finally, for a camera cluster \(\{c_0, c_1, c_2, c_3, \cdot\cdot\cdot\}\) over a road, for each \(c_i\) we process the video streams to get information as: \(c_i = \{vehicleID:[class,  direction, speed], \cdot\cdot\cdot\}\) (figure \ref{fig:Googleref}(d)). This processed data is then used to identify traffic estimates across the street network.

\section{Traffic Prediction Over Street Network}
Traffic cameras are installed over the street at specified distances on important locations like junctions and roundabouts. For example, in Brixton Road London the cameras are placed at an average of 0.4 -0.7 km apart from each other. The traffic cameras have a limited FoV extending over a few metres of the road. So, the proposed system will perform vehicle detection and tracking within this camera FoV. There is no traffic-related information available for a road segment located between the installed cameras. Thus, traffic interpolation is required to identify the unknown traffic values along with street networks.

\begin{figure}
\centering\includegraphics[width=10cm,height=5.5cm]{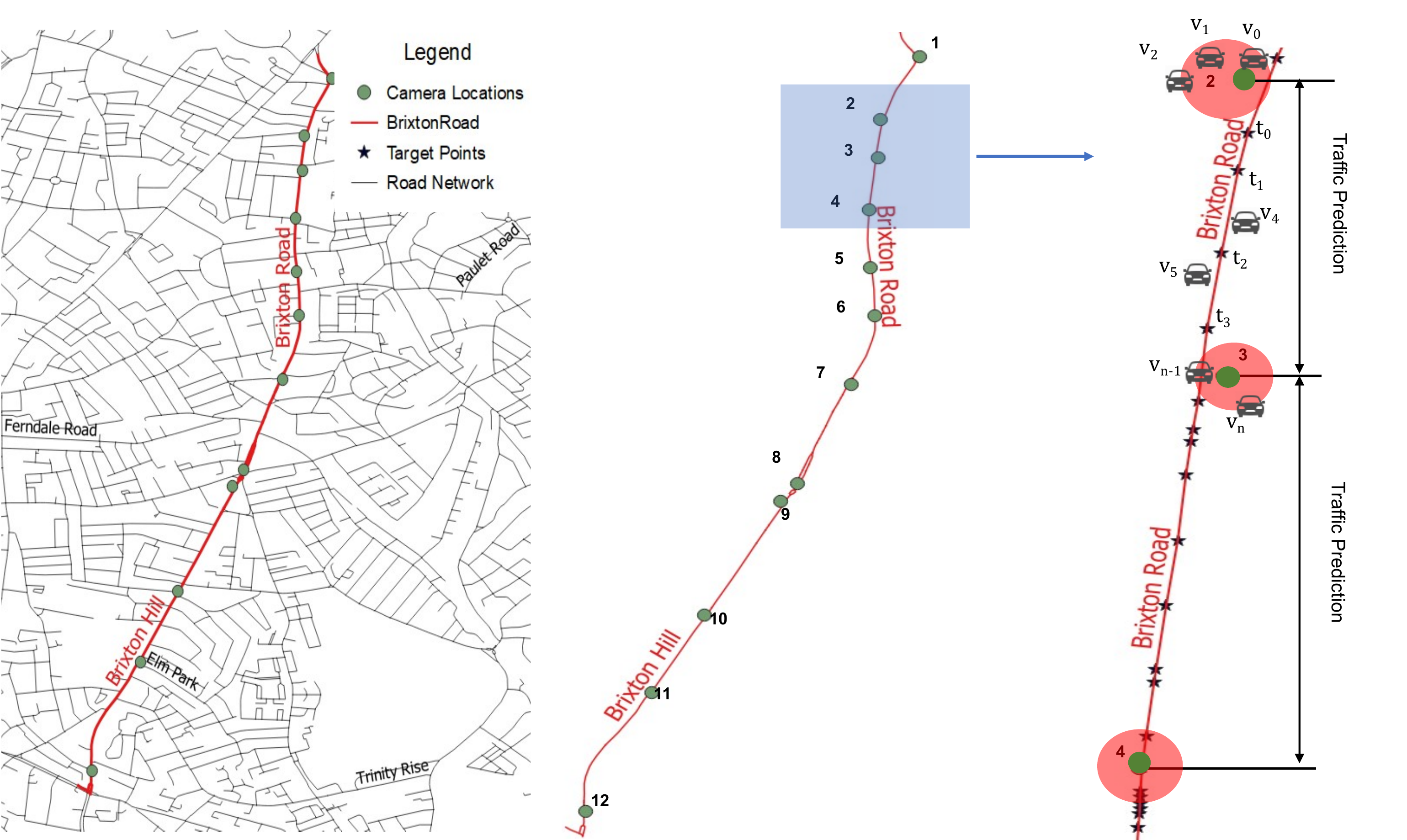}
\caption{(L-R) a)OpenStreetMap portion of London, b)Brixton Road with Camera Points, c) NB-IDW Technique Shown for Camera 2 and 3}
\label{fig:9}
\end{figure}

\begin{algorithm}[H]
\caption{Traffic Congestion Estimation Operator Algorithm}
\label{algo:1}
 \SetAlgoLined
\KwData{$Camera_{Route} \gets \{C_0, C_1, C_2, ... , C_n\}$}
\KwResult{Traffic Congestion Overlay on OSM}
\ForEach{$(C_i, C_{i+1}) \in Camera_{Route}$}
{
      $R_{ShortestPath} \gets getShortestPath(C_i, C_{i+1})$\;
      $TargetPoints \gets identifyVertices(R_{ShortestPath})$\;
      $\{O_0, O_1, O_2, ... , O_n\} \gets getObjectAndDirection(C_i, C_{i+1})$\;
      $\{S_0, S_1, S_2, ... , S_n\} \gets estimateObjectSpeed(O_i, O_1, ..., O_n)$\;
      $SamplePoints \gets \{O_0, O_1, O_2, ... , O_n\}$\;
      $dis_{network} \gets getNetworkDistance(TargetPoints)$\;
      $weights_{speed} \gets \{S_0, S_1, S_2, ... , S_n\}$\;
      $traffic_{congestion} \gets doIDW(dis_{network}, weights_{speed})$\;
      $OSM \gets displayCongestion(R_{ShortestPath}, traffic_{congestion})$\;
}
\end{algorithm}

\subsection{Network-based Traffic Interpolation}
In GIS, interpolation is widely used in mapping terrain, temperature, and pollution levels. Spatial interpolation is a prediction technique to estimate unknown values at different points using values from sample locations. Common strategies for spatial interpolation assumes that points are distributed over a 2D- Euclidean space and that the output is to be a surface spanning the entire space. Measurement of distance between sample and target locations are also done ‘as the crow flies’ along straight lines assuming that the space is isotropic and can be traversed easily in every direction. These assumptions do not hold in the case of road networks which are essentially 1D and can only be traversed along its length. Traditional spatial interpolation approaches, when applied to network-based structures results in significant errors. We performed a network-based Inverse Distance Weighted (NB-IDW) interpolation method for traffic prediction over the road segment between installed cameras ~\cite{shiode2011street}.  As per Tobler’s Law which underpins spatial interpolation techniques, we model congestion by assuming that nearby cameras that record high vehicle counts and low speed represent areas of high traffic and assume an exponential distance decay function for interpolation.

Fig. \ref{fig:9}(a) shows the Brixton Road instance in London over the OSM layer. The road stretch is approximately 2.5 miles (Camberwell New Rd/Brixton Rd to Brixton Hill/Morrish Rd) with 12 CCTV cameras (Fig. \ref{fig:9}(b)) with an average density of 4.8 cameras per mile. Fig. \ref{fig:9}(c) shows the road segment between camera 2 and 4 and how the traffic values were interpolated across this  road section. Let $\{v_0, v_1, v_2, \dots, v_{n-1}, v_n\}$ be the identified vehicles with speed $\{S_0, S_1, S_2, \dots, S_{n-1}, S_n\}$ in a specific direction for camera 2 and 3. Let $\{t_0, t_1,t_2, \dots, t_{l-1}, t_l\}$ the target locations between camera 2 and 3 with unknown traffic congestion values $\{\hat{c_0}, \hat{c_1}, \hat{c_2}, \dots, \hat{c_n-1}, \hat{c_n}\}$. Therefore, the observed traffic congestion value ($\hat{c_{o}}$) is the weighted mean of the speed of nearby identified vehicles as:
\begin{equation}
    \hat{c_{o}} = \sum_{i=1}^{n} d_{i}S_{i}
    \label{eq:3}
\end{equation}
In eq. \ref{eq:3} $d_i$ is the network distance and $S_i$ is the speed of the vehicles from the target location. The equation can be expanded as:
\begin{equation}
    \dfrac{\sum_{i=1}^{n} f(dis_{network}(v_i, t_0))S_i}{\sum_{i=1}^{n} f(dis_{network}(v_i, t_0))}
    \label{eq:4}
\end{equation}
In eq. \ref{eq:4},  $f(dis_{network} (v_i,t_0 ))$ is the network distance between the vehicles sample points and a given target location ($t_0$). The road network is treated as a graph with nodes $V$ and edges $E$. The $dis_{network}$ is the shortest path distance between the vehicles and target point and is calculated using Dijkstra algorithm while Haversine distance is used to calculate the distance. In IDW, the weights are inversely proportional to the distance and are raised to power value $p$ (-ve for inverse relationship). With the increase in $p$, the weights of farther points are decreased rapidly (eq. \ref{eq:5}). Algorithm \ref{algo:1} details the traffic congestion operator and the steps required to estimate the traffic.
\begin{equation}
    \dfrac{\sum_{i=1}^{n} (f(dis_{network}(v_i, t_0)))^{^{-p}}S_i}{\sum_{i=1}^{n} (f(dis_{network}(v_i, t_0)))^{^{-p}}}
    \label{eq:5}
\end{equation}

The traffic congestion value will be higher if vehicles are nearer to the target points. For example, in Fig. \ref{fig:9}(c) traffic congestion value at the target point $t_0$ will be more as compared to $t_1$ and $t_2$ as number of vehicles $(v_0,v_1,v_2)$ near to $t_0$ is greater. Now the question arises: how many vehicles (such as $v_3,v_4$ in Fig. \ref{fig:9}(c)) are already present in the road segment? The OSM layer for London provides the maximum speed $(S_{max})$ of the road (48 mph), so the maximum travel time between camera points can be derived. The speed of the vehicles will lie in the range of $0 \leq S_i \leq S_{max}$. Suppose the camera feeds refresh every $t_{cam-refresh}$ seconds, then the distance covered by vehicle$(v_i)$ will be:
\begin{equation}
    D_i=S_i \times t_{cam-refresh}
    \label{eq:6}
\end{equation}

Processing the camera video feeds are computationally intensive and leads to latency. The given time needs to be added to the current time to know the location of the vehicle. So, the total distance covered by the vehicle will be:
\begin{equation}
    D_i = S_i \times (t_{Cam-refresh} + processing-latency)
    \label{eq:7}
\end{equation}

If the vehicle$(v_i)$ covers the distance $D_i$ such that $0 \leq D_i \leq max_{road-length}$ where $max_{road-length}$ is the distance between two cameras, then this vehicle will be considered as sample points for traffic calculation. For example, let the distance between camera 2 and 3 is 1 mile (for easier calculation) and vehicle  $v_0$ at camera 2 is moving with speed$(S_i)$ of 40mph which is less than $S_{max}$ (48mph). Suppose the camera feed refreshes $(t_{cam-refresh})$ every 10 seconds and requires 2 seconds to process $(processing-latency)$ the new video feed. In such time the $v_0$ will cover distance$(D_i)$ = (40/3600)*(10+2) i.e. 0.13 mile and will be near point $t_0, t_1$. Thus, we consider the vehicles which are identified from both the cameras and the vehicles which are present in the road segments identified from previous feeds to perform the an appropriate traffic calculation. 

\section{Experimental Results}

\subsection{Dataset and Implementation Details}

\textbf{Traffic Camera Data}: For the experiments, two trunk roads (Brixton and Kennington) were selected from the Lambeth borough of London. As mentioned in Section 4, the traffic camera data feed was obtained from the TfL API by passing the camera location and search radius. Table \ref{table:cameralist} shows the list of cameras (total 22) installed on selected roads. A total 3080 video clips with 140 video clips (9 seconds) per camera are processed to identify the traffic status.

\begin{table}[ht]
\begin{minipage}[b]{0.45\linewidth}
\centering
\begin{adjustbox}{width=7.3cm, height=2.3cm,center}
\begin{tabular}{ | l | r | r | r |}
    \hline
    \textbf{No.} & \textbf{Brixton Road Cameras} & \textbf{Kennington Road Cameras} \\ \hline \hline
    C1 & \textcolor{blue}{\href{http://archive.tfljamcams.net/archive/Camberwell_New_Rd_Brixton_Rd/}{Camberwell New Rd/Brixton Rd}} & \textcolor{blue}{\href{http://archive.tfljamcams.net/archive/Kennington_Lane_Newington_Butts}{Kennington Lane/Newington Butts }} \\ \hline
    C2 & \textcolor{blue}{\href{http://archive.tfljamcams.net/archive/Brixton_Rd_Island_Place}{Brixton Rd/Island Place }} & \textcolor{blue}{\href{http://archive.tfljamcams.net/archive/Kennington_Pk_Rd_Penton_Pl}{Kennington Pk Rd/Penton Pl}} \\ \hline
    C3 & \textcolor{blue}{\href{http://archive.tfljamcams.net/archive/A23_Brixton_Rd_Vassell_Rd}{A23 Brixton Rd/Vassell Rd}} & \textcolor{blue}{\href{http://archive.tfljamcams.net/archive/Kennington_Pk_Rd_Braganza_St}{Kennington Pk Rd/Braganza St }}\\ \hline
    C4 & \textcolor{blue}{\href{http://archive.tfljamcams.net/archive/A23_Brixton_Rd_Hillyard_St}{A23 Brixton Rd/Hillyard St}} & \textcolor{blue}{\href{http://archive.tfljamcams.net/archive/Kennington_Pk_Rd_Kennington_Rd}{Kennington Pk Rd/Kennington Rd}}\\ \hline
    C5 & \textcolor{blue}{\href{http://archive.tfljamcams.net/archive/A23_Brixton_Rd_Ingleton_St}{A23 Brixton Rd/Ingleton St }} & \textcolor{blue}{\href{http://archive.tfljamcams.net/archive/Kenington_Pk_Rd_Kennington_Oval}{Kenington Pk Rd/Kennington Oval}}\\ \hline
    C6 & \textcolor{blue}{\href{http://archive.tfljamcams.net/archive/A23_Brixton_Rd_Wynne_Rd}{A23 Brixton Rd/Wynne Rd}} & \textcolor{blue}{\href{http://archive.tfljamcams.net/archive/A3_Clapham_Rd_Elias_Place}{A3 Clapham Rd/Elias Place}}\\ \hline
    C7 & \textcolor{blue}{\href{http://archive.tfljamcams.net/archive/Brixton_Rd_Stockwell_Pk}{Brixton Rd/Stockwell Pk }}& \textcolor{blue}{\href{http://archive.tfljamcams.net/archive/A3_Clapham_Rd_Handforth_Street}{A3 Clapham Rd/Handforth Street}}\\ \hline
    C8 & \textcolor{blue}{\href{http://archive.tfljamcams.net/archive/Acre_Lane_Coldharbour_Lane}{Acre Lane/Coldharbour Lane}} & \textcolor{blue}{\href{http://archive.tfljamcams.net/archive/A3_Clapham_Rd_Crewdson_Rd}{A3 Clapham Rd/Crewdson Rd }}\\ \hline
    C9 & \textcolor{blue}{\href{http://archive.tfljamcams.net/archive/A23_Brixton_Hill_Effra_Rd}{A23 Brixton Hill/Effra Rd }} & \textcolor{blue}{\href{http://archive.tfljamcams.net/archive/A3_Clapham_Rd_Caldwell_St}{A3 Clapham Rd/Caldwell St }}\\ \hline
    C10 & \textcolor{blue}{\href{http://archive.tfljamcams.net/archive/Brixton_Hill__Lambert_Rd}{Brixton Hill /Lambert Rd}} & \textcolor{blue}{\href{http://archive.tfljamcams.net/archive/A3_Clapham_Rd_Stockwell_Rd_Landsdowne_Way}{A3 Clapham Rd/Landsdowne Way}}\\ \hline
    C11 & \textcolor{blue}{\href{http://archive.tfljamcams.net/archive/Brixton_Hill___Elm_Park}{Brixton Hill / Elm Park }} & \textcolor{blue}{\href{http://archive.tfljamcams.net/archive/A3_Clapham_Rd_Stockwell_Park_Rd}{A3 Clapham Rd/}}\\ \hline
    C12 & \textcolor{blue}{\href{http://archive.tfljamcams.net/archive/Brixton_Hill_Morrish_Rd}{Brixton Hill/Morrish Rd }} & NA\\ \hline
   \end{tabular}
   \end{adjustbox}
    \caption{List of Traffic Cameras for Study}
    \label{table:cameralist}
\end{minipage}\hfill
\begin{minipage}[b]{0.506\linewidth}
\centering
\includegraphics[height=5cm, width=7.1cm]{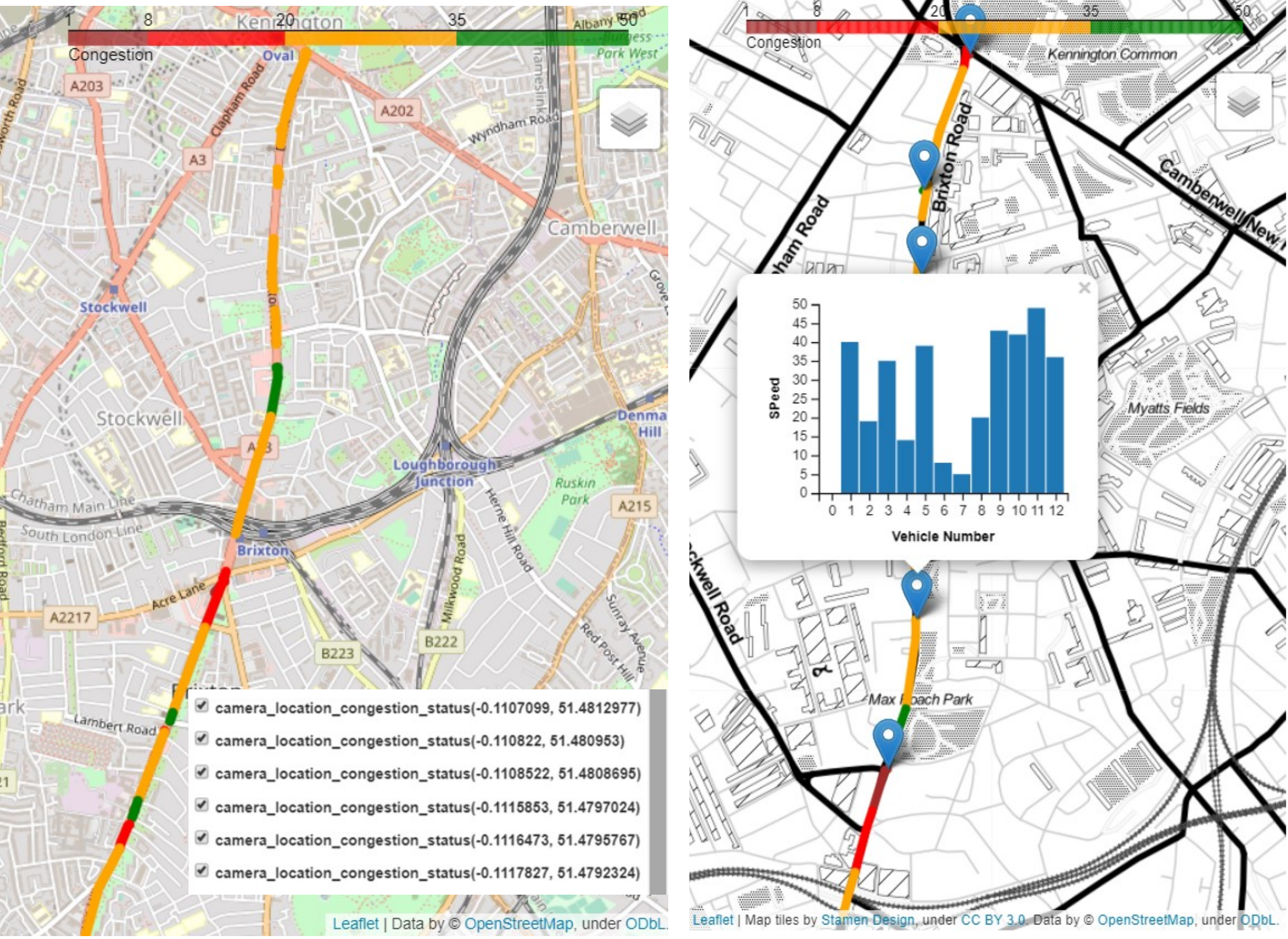}
\captionof{figure}{Traffic Congestion and Vehicle Speed Information over OSM}
\label{fig:image}
\end{minipage}
\end{table}

\noindent \textbf{Hardware and Software}: The system\footnote{\url{https://github.com/piyushy1/OSMTrafficEstimation}} was implemented in Python 3 over the VidCEP engine ~\cite{yadav2019vidcep} running on a 16 core Linux Machine with 3.1 GHz processor, 32 GB RAM and Nvidia Titan Xp GPU. The microservices were wrapped in Docker containers with Redis Stream acting as a messaging service among the containers.  The \textit{OpenCV} library was used for image processing and \textit{Darknet} and \textit{PyTorch}\footnote{\url{https://pytorch.org/}} deep learning framework were used for object detection and tracking. \textit{GStreamer} was used to stream camera video feeds from the TfL API. \textit{OSMNX} library fetched road network from OSM and calculated network distance~\cite{boeing2017osmnx}.

\noindent \textbf{Creation and Updating of Traffic Overlay over OSM}: We used the Leaflet\footnote{\url{https://leafletjs.com/index.html}} JavaScript library to overlay traffic information over OSM. The Node.js server supporting the Leaflet application was connected to the back-end system via Redis Streams. The Leaflet ColorLine class was used to highlight the roads with the given colour intensity.

\subsection{Empirical Evaluation}

\noindent \textbf{Traffic Congestion Visualization over OSM}: The traffic congestion is the interpolated value superimposed over the road segment between the cameras. As discussed in Section 6.1, the number of vehicles, their direction (left or right lane) and speed is calculated. The congestion is estimated for the road segment between the two cameras using e.q \ref{eq:5}. The value of $p$ =2 is used as the distance decay factor for interpolation. The TfL API upload the video feeds of 9 seconds (approx. 280 frames), and thus we process them in two TIME WINDOW of 4.5 seconds (140 frames) each. Using eq. \ref{eq:7}, for the second time window, we also estimate if any vehicle from the previous window is present on the road segment and take this into consideration for congestion.

Different map services calculate traffic congestion values, but they remain opaque about their method. For example, \textit{Here Maps API} provides \textit{Jam factor} values and divides them into four jam categories without explaining how the values are calculated. In this work, count and speed are considered as parameters for congestion estimation; so if the speed of vehicles is high then the traffic is considered smooth. As per OSM data, the maximum speed limit for the selected roads is 48mph (approx. 77km/hr). In Fig. \ref{fig:image}, a four-step colour-based traffic congestion over Brixton Road is shown where- 1)Green - No traffic (45-70 km/hr) 2) Orange - light traffic (30-45 km/hr), 3)Red - moderate traffic(20-30 km/hr) and 4) Brown- high traffic(10-20 km/hr). The above-defined congestion parameters range is not static and can be reconfigured depending on requirements. Fig. \ref{fig:image}(right) also shows specific traffic data available by clicking anywhere along the road. Since we could not perform a direct comparison with other map services, we can only visually compare that the traffic dynamics they provide are similar to our proposed technique.

\begin{figure}
\centering
\begin{minipage}[t]{1\linewidth}
  \centering
  \includegraphics[height=5.8cm, width=14.0cm]{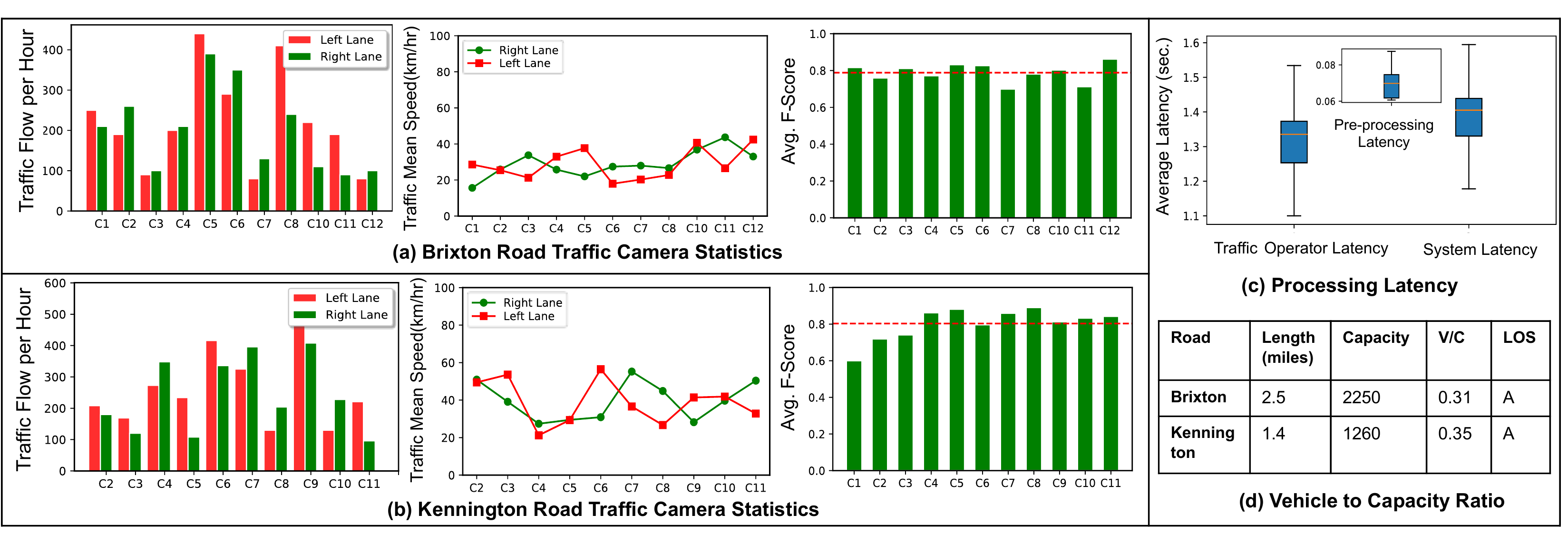}
  \captionof{figure}{Different Traffic Statistics for Roads Calculated Using Camera Video Feeds}
  \label{fig:allcam}
\end{minipage}
\end{figure}

\noindent \textbf{Traffic Flow Rate and Speed}: As per Table 1, the Traffic Flow rate (TF) is the number of vehicles in each lane per unit of time. TF is calculated for small periods and then extrapolated to larger time scales. For example, if 20 vehicles observed in 10 minutes, then TF will be 120 vehicles per hour ~\cite{mitflow}. Fig. \ref{fig:allcam}(a) and (b) shows the hourly TF for Brixton and Kennington roads for both lanes. The camera C5 (Brixton-left lane) and C9 (Kennington-left lane) have max. flow rate of 410 and 572 vehicles per hour. The reason behind such a low flow rate is the lockdown measures currently in place in London because of the COVID-19 pandemic situation. Fig. \ref{fig:allcam}(a) and (b) shows the Traffic Mean Speed (TMS) which is the average speed of vehicles recorded at a given point for a selected period. The Brixton road TMS ranges between 18-46 km/hr while that of Kennington is from 20-58 km/hr. Cameras where speed is below 20 km/hr is due to the video feeds having more red light signals. Fig. \ref{fig:image} shows the TMS example on the OSM map where markers can be clicked to get the current traffic status (e.g vehicle speed and count).

\noindent \textbf{Traffic Density and Volume to Capacity Ratio(V/C)}: Traffic density is the number of vehicles occupying a unit length of road. The Road Task Force from TfL specifies the capacity of the road as 900 vehicles per mile. Fig. \ref{fig:allcam}(d) shows the V/C ratio for both roads to identify the Level of Service (LOS). During the experimentation time, the V/C ratio for Brixton and Kennington was 0.31 (LOS-A) and 0.35 (LOS-A) respectively. As discussed earlier the lower V/C ratio is due to lockdown measures as a result of COVID-19 restrictions.

\noindent \textbf{Traffic Prediction Accuracy and Latency}: The above-discussed traffic estimates are linked to the performance of the object detector (YOLO) and tracker (DeepSORT) models. The error in these models will directly propagate to the traffic estimates and skew the overall results. F-score is the standard metric to identify the performance of the classifier. It is the harmonic mean of precision and recall and is calculated as:
\begin{equation}
    F= \dfrac{2*(Precision*Recall)}{Precision+Recall}
    \label{eq:8}
\end{equation}
In eq. \ref{eq:8}, the precision  is  the  ratio  of \textit{relevant events matched} and \textit{matched events} while recall is the ratio of \textit{relevant events matched} and \textit{relevant events}. The mean F-score of each road camera $(c_i)$ is calculated as $F_{mean} = \dfrac{\sum_{i=1}^{n}F_{ci}}{n}$. A sample of 110 video clips (each of 9 seconds) from 22 camera clusters were taken to identify the F-score. Fig. \ref{fig:allcam}(a,b) shows that the mean F-score of the Brixton and Kennington camera is 0.78 and 0.80 respectively. The error estimation rate which propagates can be calculated as ER = (actual-approx)/actual*100 and is 22\% and 20\%  respectively for both roads camera clusters. The low F-scores for some cameras were due to blurred FoV (like Kennington C1), tree shadows (like Kennington C1) and faulty cameras (e.g. Brixton C9 which was not included).

Latency measures the time required by the framework to process the traffic information and update it on the OSM. The latency can be divided into two stages: 1) Pre-processing Latency- the time required by DNN models to process the video stream to track and extract objects and 2) Traffic Operator Latency- the time required to compute traffic statistics from pre-processed data and update it on the OSM. Fig. \ref{fig:allcam}(c) shows the box plot of pre-processing and traffic and system median latency of 0.071 seconds, 1.36 and 1.42 seconds, respectively which implies a near-real-time performance.

\section{Discussion of Limitations }
While using camera feeds for traffic estimation poses privacy risks, the cameras are state infrastructure with publicly available video streams. Open-source software built on top of the open data infrastructure is more transparent compared to the opaque black-boxes and data sources which certain corporations have exclusive access. As the quality of OSM data improves, providing value-added services such as traffic estimation is essential for the adoption of OSM as a mainstream routing service. 

The uniqueness of this work lies in the query-based approach (VEQL) where users can query multiple video streams by deploying operators (e.g. traffic congestion, flow rate, speed, etc.) to the system. The framework can be deployed over the cloud and can be exposed as an API to provide services over OSM. More complex and potential traffic services like Vehicle Overtake and Lane Change ~\cite{yadav2019state} can be queried by creating operators and deploying them to the system. Our method relies on existing cameras installed along road networks and most of the limitations arise from camera deployment. For example, some cameras have sub-optimal FoVs because of obstacles. Cameras in London have a relatively low refresh rate, in the order of several minutes and provides only 9-second clips in each refresh cycle. Most of the CCTV's are installed in well-lit areas so the system can work in the night. Further, incremental weather may result in object mismatch due to lack of training datasets in such conditions. The inability to detect vehicles result in errors that propagate to final traffic computation. We have tested our model on cameras placed over straight roads and the strategy can be generalized for branched road networks. 

In terms of generating visualizations for display over maps, it is possible to use more complex spatial interpolation (e.g. kriging) and classification techniques to get better congestion estimates. Using all cameras in a city will significantly improve the sample size and provide better performance in the interpolation step leading to more accurate estimates of travel time. The various measures of traffic we offer can also be used to create comprehensive dashboards for communicating multi-dimensional aspects of traffic state. Finally, other open data streams (e.g. General Transit Feed Specification) can also be integrated to supplement traffic state calculations and estimation of travel times using different transport modes.

\section{Conclusion}
We have presented a traffic estimation framework built on open video streams based on open-source deep learning technology. We have exposed the results and data through visualizations on OSM. Exploiting the data from the video streams enabled us to extract traffic-related parameters beyond simple vehicle counts. Combining these multiple parameters provides opportunities to present a multidimensional analysis of the traffic state. For example, during interpolation, we treated each vehicle as a sample point and its speed as weight, thus factoring in both vehicle density and flow in the estimation of traffic. Users can utilize these parameters to calculate their own metrics for congestion. The VEQL query empower users to create their rules and deploy them as services for traffic-related events. We aim to deploy this service across multiple cities which make their traffic camera feeds available to be able to provide a comprehensive traffic state estimation service over OSM. We hope that places that do not make their feeds available publicly will adopt our open-source framework to provide their traffic estimation API to be consumed over OSM. Widespread adoption will provide the currently lacking feature of traffic state information on OSM and will be a step towards making OSM a viable alternative to commercial digital map service providers.



\bibliography{lipics-v2019-sample-article}

\end{document}